\documentclass[sigconf]{acmart}

\usepackage{graphicx}
 
\usepackage{amsmath,amssymb,amsthm} 
\usepackage{multirow}
\usepackage{booktabs}
\usepackage{balance}
\usepackage{dsfont}
\usepackage{graphics}
\usepackage{subfigure}
\usepackage{float}
\usepackage{hyperref}

\AtBeginDocument{%
  \providecommand\BibTeX{{%
    \normalfont B\kern-0.5em{\scshape i\kern-0.25em b}\kern-0.8em\TeX}}}

\acmDOI{10.1145/1122445.1122456}

\acmConference[MM '21] {Proceedings of the 29th ACM International Conference on Multimedia}{October 20--24, 2021}{Virtual Event, China.}
\copyrightyear{2021}
\acmYear{2021}
\setcopyright{acmcopyright}\acmConference[MM '21]{Proceedings of the 29th ACM
	International Conference on Multimedia}{October 20--24, 2021}{Virtual Event, China}
\acmBooktitle{Proceedings of the 29th ACM International Conference on Multimedia
	(MM '21), October 20--24, 2021, Virtual Event, China}
\acmPrice{15.00}
\acmDOI{10.1145/3474085.3475284}
\acmISBN{978-1-4503-8651-7/21/10}

\begin{document}
\fancyhead{}
\title{VLAD-VSA: Cross-Domain Face Presentation Attack Detection \\ with Vocabulary Separation and Adaptation}

\author{\mbox{Jiong Wang$^1$ \hspace{10pt} Zhou Zhao*$^1$ \hspace{10pt} Weike Jin$^1$ \hspace{10pt} Xinyu Duan$^2$ \hspace{10pt} Zhen Lei$^3$}}
		
\author{\mbox{Baoxing Huai$^2$ \hspace{10pt} Yiling Wu$^2$ \hspace{10pt} Xiaofei He$^1$}}
\affiliation{$^1$ College of Computer Science and Technology, Zhejiang University, China}
\affiliation{$^2$ Huawei Cloud, China \hspace{10pt} $^3$ NLPR, CASIA, China}

\email{{liubinggunzu, zhaozhou, weikejin}@zju.edu.cn, {duanxinyu, huaibaoxing, wuyiling1}@huawei.com}
\email{zlei@nlpr.ia.ac.cn, xiaofeihe@cad.zju.edu.cn}

\def\authors{Jiong Wang, Zhou Zhao, Weike Jin, Xinyu Duan, Zhen Lei, Baoxing Huai, Yiling Wu, Xiaofei He}

\renewcommand{\shortauthors}{Jiong Wang, Zhou Zhao, Weike Jin et al.}




\begin{abstract}

For face presentation attack detection (PAD),  most of the spoofing cues are subtle, local image patterns (\emph{e.g.}, local image distortion, 3D mask edge and cut photo edges). 
The representations of existing PAD works with simple global pooling method, however, lose the local feature discriminability. In this paper, the VLAD aggregation method is adopted to quantize local features with visual vocabulary locally partitioning the feature space, and hence preserve the local discriminability. We further propose the vocabulary separation and adaptation method to modify VLAD for cross-domain PAD task. 
The proposed vocabulary separation method divides vocabulary into domain-shared and domain-specific visual words to cope with the diversity of live and attack faces under the cross-domain scenario.
The proposed vocabulary adaptation method imitates the maximization step of the k-means algorithm in the end-to-end training, which guarantees the visual words be close to the center of assigned local features and thus brings robust similarity measurement.
We give illustrations and extensive experiments to demonstrate the effectiveness of VLAD with the proposed vocabulary separation and adaptation method on standard cross-domain PAD benchmarks. The codes are available at https://github.com/Liubinggunzu/VLAD-VSA.
\end{abstract}




\maketitle

\section{Introduction}

The application of face recognition, such as smart-phone unlock, access control and e-wallet payment, are usually privacy-related and widely used in the daily life.  However, the face recognition systems may be easily bypassed by the various presentation attacks modes \cite{nguyen2009your, akhtar2015biometric} (\emph{e.g.}, print attack, replay attack and 3D mask attack). 
Nowadays the face images  are easily accessed by means of social softwares and internet images, which makes face presentation attack detection (PAD, a.k.a, anti-spoofing) a vital step to
ensure the reliability of face recognition systems.

\begin{figure}[t]
	
	\begin{center}
		\includegraphics[scale=0.37]{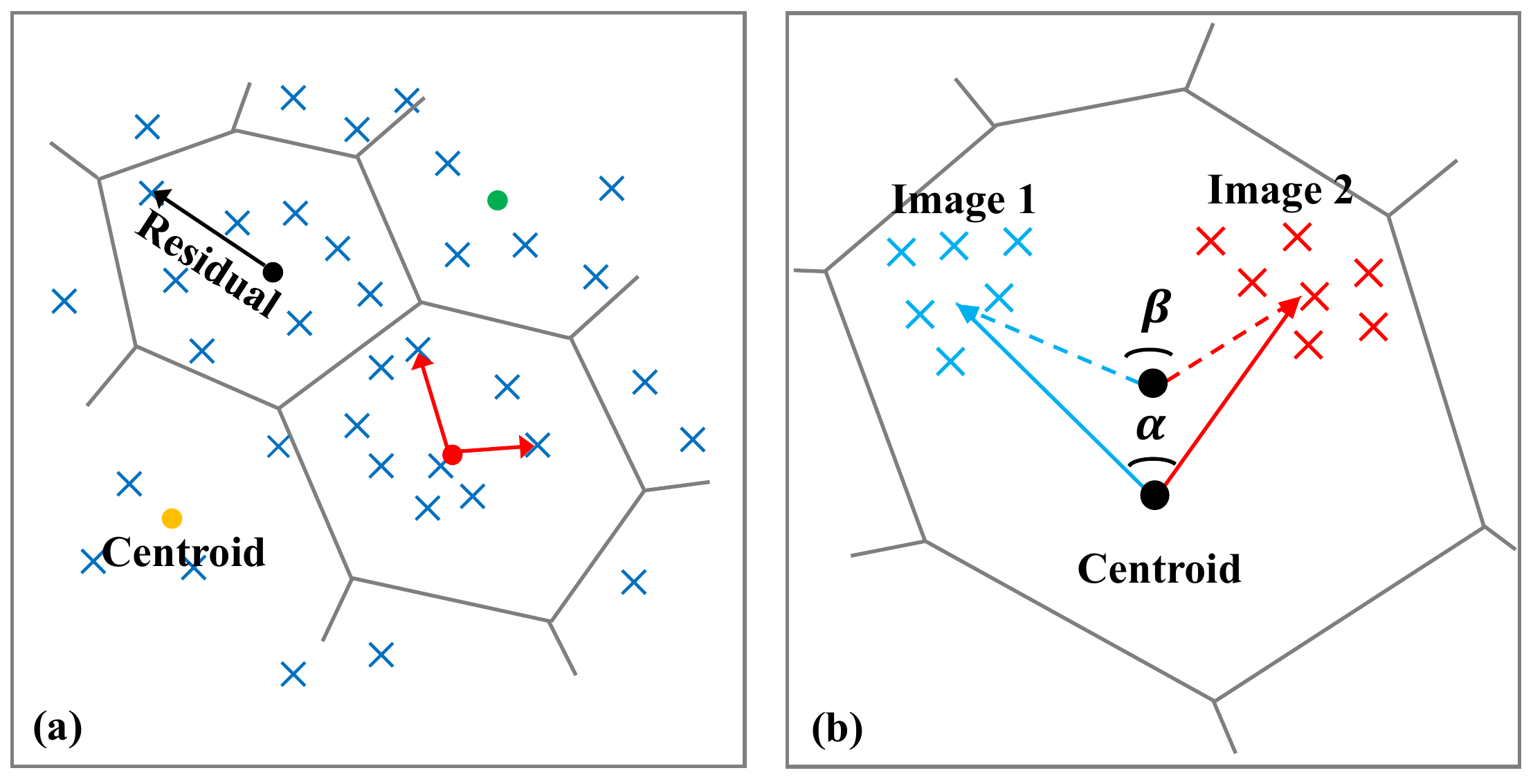}
	\end{center}
	\caption{Illustration of the local feature distribution and visual words (cluster centroids) in VLAD. (a) The visual clusters locally partition the feature space and the residuals are aggregated as representation. (b) The residual similarity is sensitive to the position of cluster centroid.
	}
	\label{fig:motivation}
\end{figure}

To tackle the face anti-spoofing (FAS) problem, numerous datasets \cite{bou17oulu, Zhang2012AFA, chingovska2012effectiveness, wen15face, tan2010face, zhang2019dataset} are released with diverse characters of subjects, attack types and modalities. Meanwhile, numerous approaches are proposed to discover the decision boundary between live and spoof faces. 

The texture-based approaches \cite{m11face, bou17face, wen15face} detect attacks by exploiting the appearance cues such as color texture and image distortion cues. The temporal-based approaches exploit temporal cues such as facial motions \cite{de2014face, shao19joint} and rPPG \cite{liu163d, liu2018learning}. Recent deep-feature-based works focus on robust PAD with auxiliary depth supervision \cite{liu2018learning, zhang2020face}, deliberate network architectures \cite{liu2019deep, yu20search, wang20deep} or exploiting the de-spoofing method \cite{jourabloo2018face, liu2020disentangling}.


Even though the above approaches report promising results in the intra-dataset test, their performance may drop significantly in the cross-dataset (domain) scenario \cite{de2013can, patel2016cross}, where the training and testing data are from different datasets with different attack types and capturing environments.
In this case, the cross-domain PAD is proposed to improve the PAD generalization ability and fit PAD models on the real world data. 
Domain adaptation techniques \cite{ganin2014unsupervised, long15learn} are introduced \cite{li2018unsupervised, wang20unsup} to align the feature distribution of source domain and target domain. 
Meanwhile domain generalization techniques \cite{motiian2017unified, li17deeper} align the feature distribution of all source domains, assuming that the resulted features are domain-invariant and thus generalize well to unseen target distributions. 
In this paper, we consider the domain generalization scenario because the target domain is usually unknown, and it is impossible to collect target data with all the variations. 


The representations of existing deep-feature-based PAD works are usually obtained by the global average pooling. We argue that this simple pooling method gives a cursory summation of the local feature maps and loses the local discriminability.
We instead adopt the VLAD aggregation method with visual clusters locally dividing the feature space and the residuals of local features to the visual words are aggregated to global representations, as illustrated in Figure~\ref{fig:motivation} (a).
Compared to conventional global pooling method, VLAD is advanced in two aspects: First, the local discriminability of features is preserved by quantizing local features to the closest visual words. 
Second, VLAD provides a reasonable selective matching kernel \cite{tolias2013aggregate} where only the intra-cluster residual similarity is compared, as discussed in section~\ref{sec:discuss}.

We further modify VLAD  with the vocabulary separation and adaptation method for cross-domain PAD. 
The vocabulary separation method tackles the diversity problem by dividing the vocabulary into domain-shared and domain-specific words. The shared words are expected to capture the attributes generalized to all domains while the specific words capture the domain-specific attributes. We propose to only align the multi-domain distribution of shared representations while setting aside the specific representations.

The proposed vocabulary adaptation method, consisting of centroid adaptation and intra-cluster discriminative loss, perfects the vocabulary optimization of VLAD.
By convention the visual words are initialized by k-means and optimized as the parameters of a convolutional layer in NetVLAD \cite{arandjelovic2016netvlad}. 
We found the optimization process is deficient because there is only the expectation step that assigns local features to visual words while no the maximization step that re-calculates the visual words as the k-means algorithm. 
As demonstrated in Figure~\ref{fig:motivation} (b), the residual similarity comparison of VLAD representation is vocabulary-sensitive. Two set of local features get a large variation of similarity under slight change of visual word. In an ideal case the visual words should localize in the assigned feature centers \cite{arandjelovic2013all}, as the maximization step of k-means, to give robust similarity measurement of VLAD representations. To this end, the proposed centroid adaptation method imitates the maximization step by calculating the center of assigned local features and expect the visual word close to the center in the mini-batch training stage. 
Furthermore, the proposed intra-cluster discriminative loss improves the intra-cluster discriminability by forcing the real and fake features far away from each other in each cluster.

In summary, 
our contributions are threefold. 
\begin{itemize}
	\item We adopt VLAD representation for cross-domain PAD task and detailedly explain, demonstrate its advantages.	
	\item We modify VLAD with the vocabulary separation (VS) and adaptation (VA) method for cross-domain PAD task. VS imposes shared-specific modeling to tackle the diversity across domains and VA perfects the vocabulary optimization of VLAD.	

	\item The extensive experiments demonstrate the effectiveness of VLAD representation with the proposed VS and VA method (VLAD-VSA), which gets new state-of-the-art result on the cross-domain PAD benchmarks. 
\end{itemize}

\begin{figure*}[t]
	
	\begin{center}
		\includegraphics[scale=0.5]{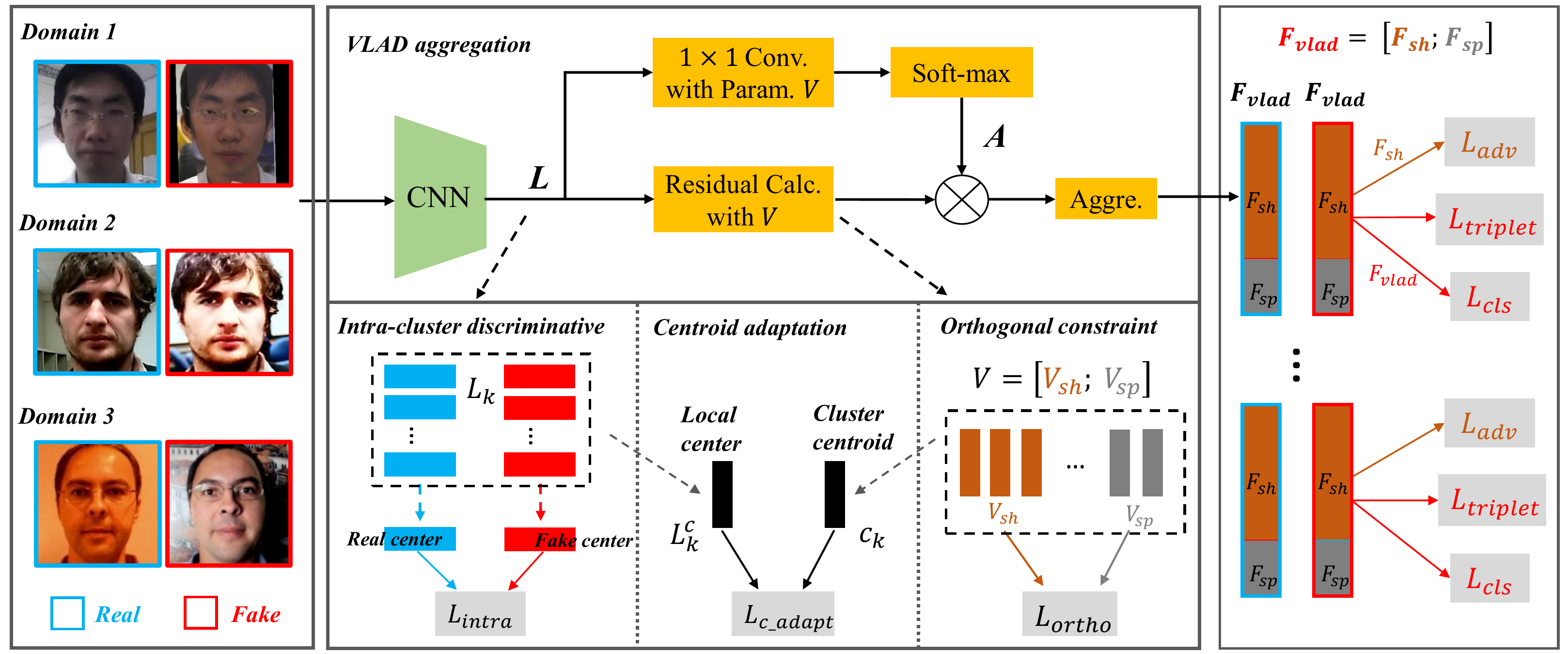}
	\end{center}
	\caption{
		Illustration of the VLAD-VSA pipeline. Real and fake faces from three domains are encoded by CNN with VLAD aggregation to get VLAD representations, which are optimized for recognition and domain alignment. The VLAD is modified by vocabulary separation (orthogonal constraint) and vocabulary adaptation(centroid adaptation, intra-cluster discriminative).
	}
	\label{fig:pipeline}
\end{figure*}
\section{Related Works}
\subsection{Face Presentation Attack Detection}
Recent PAD works can be generally grouped into intra-domain PAD and cross-domain PAD.
Conventional intra-domain PAD works aim to discover the decision boundary between live and spoof faces. 
The texture-based approaches \cite{m11face, bou17face, wen15face} detect attacks by exploiting appearance cues such as color texture and image distortion.
The temporal-based approaches exploit temporal cues such as facial motions \cite{de2014face, shao19joint} and rPPG \cite{liu163d, liu2018learning}. 
Recent deep-feature-based works focus on exploiting auxiliary supervisions \cite{liu2018learning, zhang2020face}, disentangling method \cite{jourabloo2018face, liu2020disentangling} or searching deliberate network architectures \cite{liu2019deep, yu20search, wang20deep}.
In this paper, we propose to adopt powerful VLAD representation to preserve the discriminability of local features.

Cross-domain PAD works with domain adaptation and generalization techniques aim to improve the PAD generalization ability and fit PAD models on the real world data.
For PAD in domain generalization scenario, Shao \emph{et al.} \cite{shao2019multi} propose a  multi-adversarial architecture where the feature generator is trained to compete with multiple domain discriminators simultaneously. Jia \emph{et al.} \cite{jia20single} propose a single-side adversarial learning method where the feature generator is trained to make only the real faces domain-undistinguishable, but not for the fake ones.

However, \cite{jia20single} neglects the common characters of fake images across domains and \cite{shao2019multi} ignores the individual characters of each domain. 
We admit both the common and individual characters across domains, assuming each domain contains the domain-shared characters and domain-specific characters. We argue that only the distribution of domain-shared attributes should be aligned. 
In this way, the proposed VS method divides vocabulary into domain-shared words and domain-specific words, and only the shared representation distribution of source domains is aligned in the training stage.

\subsection{Vector of Locally Aggregated Descriptors}

Bag-of-Visual-Words (BoVW) model \cite{sivic2003video} is one of the most widely used encoding methods for visual recognition before the prevalence of deep learning. Departing from BoVW, VLAD \cite{jegou2010aggregating} and Fisher vector \cite{sanchez2013image} demonstrate their effectiveness in image classification and retrieval by encoding first-order and second-order statistics. 
Combining with CNN framework, NetVLAD \cite{arandjelovic2016netvlad} implements VLAD as a trainable aggregation layer with differentiable operations for end-to-end learning.
The effectiveness of NetVLAD over conventional global pooling is proved in various tasks such as place recognition \cite{arandjelovic2016netvlad}, action recognition \cite{girdhar2017actionvlad} and video face recognition \cite{zhong2018ghostvlad}.

The vocabulary of NetVLAD is usually initialized by the k-means clustering algorithm, 
where visual words are the cluster centroids. But the vocabulary is optimized as the parameters of convolutional layer in the training stage, and there are no guarantee that the visual words localize in the cluster center to give robust similarity measurement.
The proposed vocabulary adaptation method solves this problem by guaranteeing the visual words close to the center of assigned local features in the batch-training process.

\subsection{Domain Generalization}

Domain generalization (DG) aims to learn a domain-agnostic model from multiple source domains
that can then be applied to the unseen target domain.  
Many works propose to learn domain invariant features \cite{muan13domain, li18domain} or augment the source domain to a wider data space \cite{volpi18gener, shan18gener}, enlarging the possibility of covering target domain. Meta-leaning based techniques \cite{balaji2018metareg, shao2019regularized, qin2019learning} are also exploited for DG.

The proposed vocabulary separation method shares similar motivation with the shared-specific modeling DG works \cite{daume2009frustratingly, khosla2012undoing, bou16domain}, 
which devise domain-specific and domain-agnostic encoders \cite{bou16domain, li17deeper} or classifiers \cite{khosla2012undoing, piratla2020efficient}.
In comparison, the proposed VS imposes shared-specific modeling in vocabulary-level which quantizes domain-shared and specific local features. Meanwhile,  
different from above works which discard the domain-specific components and only retain the shared components after training, we combine both the shared and specific components for recognition because the domain-specific components such as specific attack types in PAD are informative for attack detection.

\section{The Proposed Method}

In this section,  we first introduce the conventional cross-domain baseline method (section~\ref{sec:Preliminary}), then describe the NetVLAD pipeline (section~\ref{sec:netvlad}) and discuss its advantages (section~\ref{sec:discuss}). We finally describe the VLAD-VSA representation with vocabulary separation  (section~\ref{sec:vs}) and adaptation (section~\ref{sec:va}) method to modify NetVLAD for cross-domain PAD scenario.

\subsection{Preliminary}   \label{sec:Preliminary}
In cross-domain PAD, there are images $X = \{X_1, X_2, ..., X_S\}$ from $S$ source domains $D_S = \{D_1, D_2, ..., D_S\}$, with corresponding labels $Y = \{Y_1, Y_2, ..., Y_S\}$ ($Y \in \{0, 1\}$, real or fake). The aim is to learn a generalized classifier on source domains to identify real and attack images on the unseen target domain.
A conventional cross-domain PAD baseline adopts a generator $G$ with CNN and global average pooling (GAP) to get global representations, which are optimized with a classification loss, triplet loss to separate the real and fake faces in feature space for discriminability.
The generator $G$ plays a min-max game with the discriminator $D$  designed to successfully distinguish domains, forming the adversarial loss to get generalized features for all source domains.

Formally, given the local feature maps $L = \{f_i\}_{i=1}^{N}$ of final convolutional layer with size of $N \times d$, where $N$ is the total number of flattened local features with dimension $d$. Conventional GAP summarizes all local features to a global image representation as formulated:
\begin{equation} 
G(x) = F_{avg}(L)= \frac{1}{N} \sum_{i=1}^{N}f_{i}, f_{i} \in x
\end{equation}
The representations are optimized by a cross-entropy loss for classification as follow:
\begin{equation} 
\mathcal{L}_{cls}(X, Y; G) = -\mathbb{E}_{x, y \sim X,Y}  \sum\nolimits_{k=0}^{1} \mathds{1}_{[k=y]} \log G(x).
\end{equation}
The two-category triplet loss complements the classification loss by separating the real and fake faces in feature space:
\begin{equation} 
\begin{aligned}
\mathcal{L}_{triplet}(X, Y; G) &= \sum_{\forall y_a = y_p, y_a \neq y_n} (\left \| G(x_a) - G(x_p) \right \|_F^2 \\ 
&-  \left \| G(x_a) - G(x_n) \right \|^2 + m ),
\end{aligned}
\end{equation}
where $\left \| \right \|_F^2$ is the squared Frobenius norm, $m$ is the margin,  $x_a$ has same label with positive sample $x_p$ and has dissimilar label with negative sample $x_n$.
The adversarial loss aligns the distribution of all source domains and guarantees the generalization ability to unseen domain:
\begin{equation} 
\begin{aligned} \label{eq:adv}  
\min_{D} \max_{G} &\mathcal{L}_{adv}(X; D, G)  = \\ &-\mathbb{E}_{x, y \sim X,Y_D}  \sum\nolimits_{s=1}^{S} \mathds{1}_{[s=y]} \log D(G(x)),
\end{aligned}
\end{equation}
where $Y_D \in \{1, 2, ..., S\}$ denotes domain labels.

In this way, the conventional baseline method can distinguish the real and fake faces across domains with generalization and discriminative ability. We improve this baseline by introducing VLAD aggregation method and modifying VLAD with the proposed vocabulary separation and adaptation strategy.

\subsection{NetVLAD Representation} \label{sec:netvlad}

Given local features $L = \{f_i\}_{i=1}^{N}$ and vocabulary 
$V = \{c_k\}_{k=1}^{K}$ with $K$ visual clusters dividing the feature space, the local features are first quantized to the closest visual words in VLAD aggregation and the resulted assignment matrix $A \in \mathbb{R}^{N \times K}$ is formulated:
\begin{equation} 
A = LV^T, A_{i,j} = f_i c_j^{T}. 
\end{equation}

In NetVLAD \cite{arandjelovic2016netvlad}, the assignment step is implemented by an $1 \times 1$ convolutional layer with parameter $V$ on the local features, followed by the soft-max function to scale the numeric as shown in the VLAD aggregation of Figure~\ref{fig:pipeline}. 
The residuals between the local features and their assigned cluster centroids are calculated, and then they are weighted by the assignment scores and aggregated into the NetVLAD representation $F_{vlad} \in \mathbb{R}^{K \times d}$. The $k$-th NetVLAD component is formulated as follow:
\begin{equation}
F_{vlad}^{k} = \sum _{i=1}^{N} \underbrace{ \frac{e^{t f_i c_k^T}}{\sum_{{k}'} e^{t  f_i c_{{k}'}^T }} } _\text{Soft-assignment} \underbrace{ (f_i - c_k) }
_\text{Residual $r_{i, k}$},
\end{equation}
where $t$ is the temperature. For $t \rightarrow + \infty$, it corresponds to the hard-assignment situation where the assign scores of closest clusters are 1 and 0 otherwise.  In our experiment, the soft-assignment is considered to tackle the quantization noise and $t$ is empirically set to 3. The NetVLAD representation is then intra-normalized, flattened, and $l2$-normalized \cite{arandjelovic2013all, arandjelovic2016netvlad} to a global vector for recognition and domain alignment.

\subsection{Selectivity of VLAD} \label{sec:discuss}
Essentially, the similarity comparison of global representations is equivalent to the matching of local features \cite{stylianou2019visualizing}. With conventional global average pooling, the similarity comparison of representation corresponds to all-to-all matching of local features \cite{babenko2015aggregating, stylianou2019visualizing} as formulated: 
\begin{equation} 
\begin{aligned}
\left \langle F_{avg}(x_1),F_{avg}(x_2)\right \rangle &= \big \langle \sum_{f_{i} \in x_1} f_i, \sum_{f_{j}\in x_2} f_j \big \rangle \\
&= \sum_{f_i \in x_1}\sum_{f_j\in x_2} \left \langle f_i, f_j \right \rangle , 
\end{aligned}
\end{equation}
where we omit the averaging for clarity.
Analogously, the similarity comparison of VLAD representations is essentially the summation of word-level residual comparison:

\begin{equation} 
\begin{aligned}
\left \langle F_{vlad}(x_1),F_{vlad}(x_2)\right \rangle & = \sum_{{k}} \left \langle F_{vlad}^k(x_1),F_{vlad}^k(x_2)\right \rangle, \\
\end{aligned}
\end{equation}
where the similarity comparison of $k$-th word is formulated as follow:
\begin{equation} 
\begin{aligned}
\left \langle F_{vlad}^k(x_1),F_{vlad}^k(x_2)\right \rangle & = \big \langle \sum_{f_{i} \in x_1, k} r_{i,k}, \sum_{f_{j}\in x_2, k} r_{j,k} \big \rangle \\
&=  \sum_{f_i \in x_1, k} \sum_{f_j\in x_2, k} \left \langle r_{i,k}, r_{j,k} \right \rangle ,
\end{aligned}
\end{equation}
in which $r_{i,k} = f_i - c_k$ is the residual of $i$-th local feature to $k$-th cluster. We choose the hard-assignment situation and omit the assignment scores for clarity.

It can be outlined that VLAD provides a selective matching kernel \cite{tolias2013aggregate} where the local features are assigned to closest clusters and only the intra-cluster similarity of residuals is compared. Compared to the all-to-all matching kernel, selective matching kernel is reasonable for human common sense because human usually compare same facial parts when comparing two faces. 

\subsection{Vocabulary Separation} \label{sec:vs}

The unlimited variations of the dominant conditions (illumination, facial appearance, camera quality, etc.) in face acquisition bring diverse data distribution and difficulty for representation distribution alignment. This situation is more pronounced when different datasets have different attack types. 
We assume all the domains have shared components and each domain has its specific components. Both the shared and specific components are helpful for recognizing the attacks, and only the shared components need to be aligned across domains. 

The proposed vocabulary separation method divides the vocabulary into $K_1$ shared visual words $V_{sh} \in \mathbb{R}^{K_1 \times d}$ and $K_2$ specific words $V_{sp} \in \mathbb{R}^{K_2 \times d}$. The resulted shared representations $F_{sh}$ and specific representations $F_{sp}$ are got by these two kinds of words, respectively. 
In the optimization process, we combine both the shared and specific representations for recognition but only align the distribution of shared representations with the adversarial loss (Equation~\ref{eq:adv}), as illustrated in the right part of Figure~\ref{fig:pipeline}.

Following former shared-specific modeling works \cite{bou16domain, piratla2020efficient}, we constrain the shared and specific words to be orthogonal, expecting they 
capture diverse and unrelated cues: 

\begin{equation} 
\mathcal{L}_{ortho} = || V_{sh} V_{sp}^T ||_F^2 .
\end{equation}

\begin{figure}[t]
	
	\begin{center}
		\includegraphics[scale=0.37]{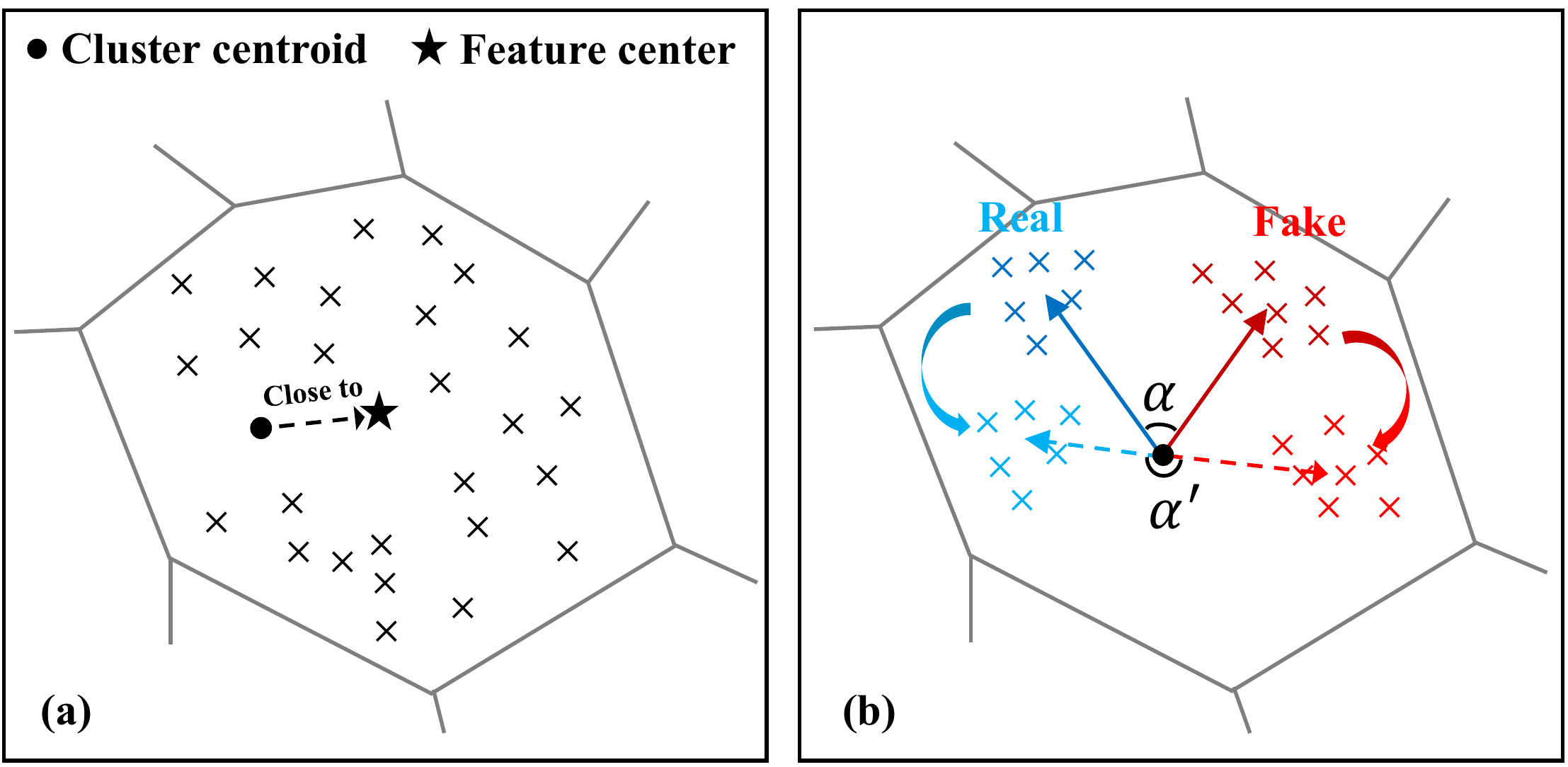}
	\end{center}
	\caption{
		Illustration of the proposed vocabulary adaptation method. In each cluster, (a) the assigned feature center is calculated and the cluster centroid is expected to be close to it. (b) The angle between real and fake residual centers is expected to be large to improve the intra-cluster discriminability.
	}
	\label{fig:motivation2}
\end{figure}

\subsection{Vocabulary Adaptation} \label{sec:va}  
The visual words of VLAD are usually initialized by k-means algorithm where the final expectation step assigns local features to closest visual words and maximization step calculates the center of assigned features as visual words. 
The training stage of NetVLAD includes only expectation step which assigns local features to visual words but without the maximization step that updates the cluster centroids.
So we propose a centroid adaptation method to imitate the maximization step by minimizing the distance between the cluster centroids and the centers of their assigned local features.

Formally, after assigning local features to their closest visual words, the assigned local feature set of $k$-th cluster (hard-assignment) are formulated as $L_k = \{f_{i,k}\}_{i=1}^{N_k}$, in which $N_k$ is the number of assigned features. As illustrated in centroid adaptation of Figure~\ref{fig:pipeline} and Figure~\ref{fig:motivation2} (a), the center of assigned features is calculated and the cluster centroid is expected to be close to it. In this way, the centroid adaptation loss is formulated as:
\begin{equation} 
\mathcal{L}_{c\_adapt} = \sum_{k=1}^{K}  || L_k^{c} - c_k ||_F^2,  L_k^{c} = \frac{1}{N_k} \sum_{i=1}^{N_k} f_{i, k}.
\end{equation}
To further improve the intra-cluster discriminability, we apply an explicit constraint in each cluster to force the feature center of real images far away from the fake images. As illustrated in Figure~\ref{fig:motivation2} (b), we force the angle $\alpha$ between the real residual center and fake residual center to be large, which forms the intra-cluster discriminative (intra) loss:  

\begin{equation} 
\mathcal{L}_{intra} = \sum_{k=1}^{K}  (1 - || r_{k}^{real_c} -  r_{k}^{fake_c} ||_F^2),
\end{equation}
where $r_{k}^{real_c}$ and $r_{k}^{fake_c}$ are respectively the residual center of real and fake faces in $k$-th cluster.

Note that we name the combination of centroid adaptation and intra loss as vocabulary adaptation, because we found combining them together gets consistent improvement while the performance improvement of single one method is not stable. We suppose the reason is these two losses are calculated within mini-batch, suffering from the sample bias with biased feature distribution.

In summary, the optimization of our VLAD-VSA representation includes the classification and triplet loss to separate real and fake faces in feature space, the adversarial loss to align the multi-domain distribution of shared representations, the orthogonal loss to constrain the shared words capture unrelated cues with the specific words, the centroid adaptation loss to guarantee the visual words localize in the center of the assigned features, the intra loss to distinguish the real and fake features inside clusters, as formulated:

\begin{equation} \label{eq:all}
\mathcal{L} = \mathcal{L}_{cls} + \lambda_1 \mathcal{L}_{triplet} + \lambda_2 \mathcal{L}_{adv} + \lambda_3 \mathcal{L}_{ortho} + \lambda_4 \mathcal{L}_{c\_adapt} + \lambda_5 \mathcal{L}_{intra} 
\end{equation}

\begin{table}[]
	\centering
	\caption{A summary of four FAS datasets (train and test split) used for evaluation. The attack type $P$ denotes printed photo, $D$ denotes display photo, $R$ denotes replayed video, $C$ denotes cut photo.}
	\begin{tabular}{c|c|c|c}
		\toprule
		\multirow{2}*{Dataset} & Attack & Capture & Real / Fake\\
		& type & Device &  (Video) \\
		\hline
		O & $P, D, R$ & Mobile camera & 720 / 2880 \\
		\hline
		\multirow{2}*{C} & \multirow{2}*{$P, C, R$} & USB camera & \multirow{2}*{150 / 450} \\
		& & Digital camera &\\
		\hline
		I & $P, D, R$ & Laptop camera & 140 / 700\\
		\hline
		\multirow{2}*{M} & \multirow{2}*{$P, R$} & Laptop camera & \multirow{2}*{70 / 210} \\
		& & Mobile camera &\\
		\bottomrule
	\end{tabular}
	\label{tab:dataset}
\end{table}
\section{Experiments}  
\subsection{Datasets and Evaluational Metric}
We evaluate the cross-domain FAS performance on four public datasets: OULU-NPU \cite{bou17oulu} (O for short), CASIA-FASD \cite{Zhang2012AFA} (C for short), Idiap Reply-Attack \cite{chingovska2012effectiveness} (I for short) and MSU-MFSD \cite{wen15face} (M for short). As summarized in Table~\ref{tab:dataset}, four datasets are collected with diverse capture devices, attack types, illumination conditions, background scenes and races. Hence significant domain shift exists  among these datasets.      

We select three datasets as source domains for training and the remaining one as target domain for evaluation. In this way, we have four evaluation settings in total: O\&C\&I to M, O\&M\&I to C, O\&C\&M to I, and I\&C\&M to O.
The Half Total Error Rate (HTER, the average of false acceptance rate and false rejection rate) and the Area Under Curve (AUC) are adopted as the evaluation metrics following previous cross-domain PAD works \cite{shao2019multi, wang20cross}.
\begin{table*}[]
	\centering
	\caption{Comparison with state-of-the-art Cross-domain face PAD works under four settings.}
	\begin{tabular}{c|c|c|c|c|c|c|c|c}
		\toprule
		\multirow{2}*{\textbf{Method}} & \multicolumn{2}{c|}{\textbf{O\&C\&I to M}} & \multicolumn{2}{c|}{\textbf{O\&M\&I to C}} & \multicolumn{2}{c|}{\textbf{O\&C\&M to I}} & \multicolumn{2}{c}{\textbf{I\&C\&M to O}} \\
		\cline{2-9}
		~ & HTER(\%) & AUC(\%) & HTER(\%) & AUC(\%) & HTER(\%) & AUC(\%) & HTER(\%) & AUC(\%)  \\
		\hline
		\hline
		MS-LBP \cite{maatta2011face}         & 29.76 & 78.50 & 54.28 & 44.98 & 50.30 & 51.64 & 50.29 & 49.31 \\
		Binary CNN \cite{yang2014learn}     & 29.25 & 82.87 & 34.88 & 71.94 & 34.47 & 65.88 & 29.61 & 77.54 \\
		IDA \cite{wen15face}               & 66.67 & 27.86 & 55.17 & 39.05 & 28.35 & 78.25 & 54.20 & 44.59 \\
		Color Texture \cite{boulkenafet2016faceLBP}         & 28.09 & 78.47 & 30.58 & 76.89 & 40.40 & 62.78 & 63.59 & 32.71 \\
		LBP-TOP \cite{de2014face}            & 36.90 & 70.80 & 42.60 & 61.05 & 49.45 & 49.54 & 53.15 & 44.09 \\
		Auxiliary (Depth) & 22.72 & 85.88 & 33.52 & 73.15 & 29.14 & 71.69 & 30.17 & 77.61 \\
		Auxiliary \cite{liu2018learning} & -     & -     & 28.40 & -     & 27.60 & -     & -     & -     \\
		MADDG (M) \cite{shao2019multi}           & 17.69 & 88.06 & 24.50 & 84.51 & 22.19 & 84.99 & 27.89 & 80.02 \\
		MD-DRL \cite{wang20cross} & 17.02 & 90.01 & \textbf{19.68} & \textbf{87.43} & 20.87 & 86.72 & 25.02 & 81.47 \\
		SSDG (M) \cite{jia20single} & 16.67 & 90.47 & 23.11 & 85.45 & 18.21 & \textbf{94.61} & 25.17 & 81.83\\
		\textbf{VLAD-VSA (M)} & \textbf{11.43} & \textbf{96.44}& 20.79& 86.32 &\textbf{12.29}& 92.95&\textbf{21.20} & \textbf{86.93}\\
		\hline
		SSDG (R) \cite{jia20single} & 7.38  & 97.17 & 10.44 & \textbf{95.94} & 11.71 & 96.59 & 15.61 & 91.54 \\
		\textbf{VLAD-VSA (R)} &\textbf{4.29} & \textbf{98.25}& \textbf{8.76}& 95.89 &  \textbf{7.79}  & \textbf{97.79}& \textbf{12.64} & \textbf{94.00}\\
		\bottomrule
	\end{tabular}
	\vspace{-0.1in}
	\label{tab:comparison_sota}
\end{table*}

\begin{table*}[]
	\centering
	\caption{Ablation of the VLAD representation, vocabulary adaptation and separation method with ResNet backbone.}
	\begin{tabular}{l|c|c|c|c|c|c|c|c}
		\toprule
		\multirow{2}*{\textbf{Method}} & \multicolumn{2}{c|}{\textbf{O\&C\&I to M}} & \multicolumn{2}{c|}{\textbf{O\&M\&I to C}} & \multicolumn{2}{c|}{\textbf{O\&C\&M to I}} & \multicolumn{2}{c}{\textbf{I\&C\&M to O}} \\
		\cline{2-9}
		~ & HTER(\%) & AUC(\%) & HTER(\%) & AUC(\%) & HTER(\%) & AUC(\%) & HTER(\%) & AUC(\%)  \\
		\hline
		GAP baseline & 8.33 & 95.80& 14.72 & 92.70 & 17.64  &  91.42  & 14.48 & 92.67\\
		VLAD  &7.14 & 97.81& 12.02 &  94.01& 15.64 &93.72& 14.20 &93.25 \\
		\hline
		VLAD + VS &5.71 & 97.86& 11.46 & 94.65 & 9.14 & 96.33&13.58  & 93.99\\
		VLAD + VA &5.71 & 97.86& 10.23&94.88 & 12.14  & 94.96& 13.16 & 94.00 \\
		VLAD-VSA &\textbf{4.29} & \textbf{98.25}& \textbf{8.76}& \textbf{95.89}&  \textbf{7.79}  & \textbf{97.79}& \textbf{12.64} & \textbf{94.00}\\
		\bottomrule
	\end{tabular}
	\label{tab:ablation}
\end{table*}

\subsection{Implementation Details}

The MTCNN algorithm is adopted for face detection and alignment, and all the detected faces are resized to 256 $\times$ 256 $\times$ 3 as network inputs. We follow the one-frame setting \cite{shao2019multi, jia20single} where one frame in one video is selected for training and two frames for testing. 
We implement the proposed method on the ResNet-18 (R) \cite{he2016deep} and MADDG (M) \cite{shao2019multi} backbone, and replace the final convolutional layer with a 128-d convolutional kernel to reduce the dimensionality of local features.  The NetVLAD layer is followed to get global image representations.
The vocabulary of VLAD is randomly initialized and the vocabulary separation is manually set because the final conv-layer is re-initialized and there are no cues for shared and specific words. 
The SGD optimizer is used with batch size of 60 (10 real and 10 fake faces for each domain), initial learning rate is 0.001, which is dropped to 0.0001 after 1500 iterations.  In most cases, the $\lambda$ in Equation~\ref{eq:all} is set to be 0.1, and there are slightly difference for the four settings. We recommend the readers to refer to the released codes.
Note that most of the ablation studies are implemented on ResNet-18 architecture because of its higher performance and lower flops, memory usage. 

\subsection{Comparison with State-of-the-Arts}
As shown in Table~\ref{tab:comparison_sota}, the cross-domain PAD performance of several former representative face anti-spoofing methods, such as Multi-Scale LBP (MS-LBP) \cite{maatta2011face}; Binary CNN \cite{yang2014learn}; Image Distortion Analysis (IDA) \cite{wen15face}; Color Texture (CT) \cite{boulkenafet2016faceLBP}; LBPTOP \cite{de2014face} and Auxiliary supervision \cite{liu18learn} are listed for comparison. 
The performance of recent state-of-the-art cross-domain face anti-spoofing works such as MADDG \cite{shao2019multi}, SSDG \cite{jia20single} and MD-DRL \cite{wang20cross} are also shown.
Compared with recent works with well-designed triplet loss and elaborate adversarial loss \cite{shao2019multi, wang20cross}, we adopt naive two-category triplet loss and adversarial loss, without introducing the additional depth, ID supervision as \cite{shao2019multi, wang20cross}. The proposed VLAD-VSA method is advanced in adopting VLAD representation and modifying NetVLAD with vocabulary separation and adaptation.

It can be seen that VLAD-VSA significantly surpasses most of existing works under the four cross-domain evaluation setting with the MADDG (M) and ResNet (R) backbones. The performance gaps are more pronounced for the weaker MADDG backbone, which demonstrates VLAD representations with the proposed vocabulary adaptation and separation strategies have excellent generalization and discriminative ability for cross-domain PAD task.
The performance of VLAD-VSA is slightly worse than the MD-DRL \cite{wang20cross} on the O\&M\&I to C setting with MADDG backbone. We suppose the reason is MD-DRL adopts auxiliary ID supervision and devises three encoders with three-times more parameters. Even so, we still  outperform it by large margins under the remaining three settings with two backbones.

\begin{table*}[]
	\centering
	\caption{Ablation of VLAD performance with the proposed vocabulary adaptation and separation method (ResNet).}
	\begin{tabular}{l|c|c|c|c|c|c|c|c}
		\toprule
		\multirow{2}*{\textbf{Method}} & \multicolumn{2}{c|}{\textbf{O\&C\&I to M}} & \multicolumn{2}{c|}{\textbf{O\&M\&I to C}} & \multicolumn{2}{c|}{\textbf{O\&C\&M to I}} & \multicolumn{2}{c}{\textbf{I\&C\&M to O}} \\
		\cline{2-9}
		~ & HTER(\%) & AUC(\%) & HTER(\%) & AUC(\%) & HTER(\%) & AUC(\%) & HTER(\%) & AUC(\%)  \\
		\hline
		VLAD  & 7.14 &97.81 & 12.02 &94.01 & 15.64 &93.72 & 14.20 &93.25 \\
		+ VS w/o specific & 5.95 & 98.08 & 14.05 & 92.06 & 14.29 & 94.33 & 14.05 & 93.07 \\
		+ VS w/o ortho & 7.14  &  97.27&  13.26 & 93.38 & 10.79 & 95.17 &  15.28 & 92.49  \\
		+ VS &  \textbf{5.71} & \textbf{97.86} & \textbf{11.46} & \textbf{94.65} & \textbf{9.14} & \textbf{96.33} & \textbf{13.58}  & \textbf{93.99} \\
		\hline
		VLAD & 7.14 &97.81 & 12.02 &94.01 & 15.64 &93.72 & 14.20 &93.25\\
		+ VA w/o c$\_$adapt & 7.38 & 97.99 &  11.46 & 94.50 & 17.93 & 90.95 & 14.44 &  93.27 \\
		+ VA w/o intra &8.33 & 97.32&  10.58  & 94.88 & 17.00 & 91.65& 13.61 & 93.48  \\
		+ VA & \textbf{5.71} & \textbf{97.86} & \textbf{10.23} & \textbf{94.88} &  \textbf{12.14}  & \textbf{94.96} & \textbf{13.16} & \textbf{94.00} \\
		\bottomrule
	\end{tabular}
	\label{tab:ablation2}
\end{table*}

\subsection{Ablation Study}
In this subsection we give exhausted ablation studies to demonstrate the effectiveness of VLAD representation, vocabulary separation (VS) and adaptation (VA). 
In Table~\ref{tab:ablation}, the results of conventional cross-domain PAD baseline (section~\ref{sec:Preliminary}) are presented. When substituting the GAP with VLAD aggregation, there is consistent performance increase for all settings and metrics. 
When coupled with the proposed VS method, imposing shared-specific modeling, VLAD is improved with preferable performance. Likewise, the proposed VA method, consisting of centroid adaptation loss and intra-cluster discriminative loss, improves VLAD representation consistently. We can also observe the performance of VLAD is further boosted when simultaneously coupled with VS and VA (VLAD-VSA).

We also give more detailed ablations of VS and VA method in Table~\ref{tab:ablation2}. 
\textbf{VS. } In the former shared-specific modeling works \cite{bou16domain, li17deeper, piratla2020efficient}, the specific components are discarded and regarded useless for recognition. For the proposed VS method, we argue that the specific features are useful for PAD recognition.  In Table~\ref{tab:ablation2}, we compare with the conventional shared-specific modeling (VS w/o specific). 
We can see that disabling the specific features gets worse results for all settings and metrics compared to the proposed VS method, which confirms our assumption that the specific components describe the diverse attack types or other specific but beneficial patterns. 
The orthogonal constraint is applied on the shared and specific words to expect they capture diverse cues. 
We can see that orthogonal constraint is necessary for vocabulary separation and disabling it ((VS w/o ortho) gets worse results. We also found the orthogonal constraint accelerates the convergence in the training stage.

\textbf{VA.} The VA method consists of centroid adaptation and intra-cluster discriminative loss. In Table~\ref{tab:ablation2}, we can observe solely centroid adaptation (VA w/o intra) or intra-cluster discriminative loss (VA w/o c$\_$adapt) may not consistently improve the performance. The performance is slightly degraded under O\&C\&I to M and O\&C\&M to I settings. But these two losses are complementary each other and bring consistent improvements when combined together. We suppose the reason is these two losses suffer from sampling bias in the mini-batch training strategy. 
When combined together, the c$\_$adapt loss stables intra loss by giving reasonable similarity measurement, and intra loss consolidates c$\_$adapt loss with a reasonable distribution of real and fake features. The effectiveness of these two loss is visually demonstrated  in section~\ref{sec:quali}.

\textbf{Shared-specific modeling.}
The proposed VS method divides the vocabulary into $K_1$ domain-shared and $K_2$ specific visual words, and only aligns the multi-domain distribution of shared representations.
We investigate the influence of $K_2$ choice in Table~\ref{tab:sp_num}, in which the conventional VLAD can be viewed as a special case of VS when $K_2=0$ ($K_1=32$), assuming all the clusters are shared across domain. 
It can be observed that VS method with moderate $K_2$ choice enhances the VLAD performance. But we can observe performance drop when adopting $K_2 = 6$ or 10, compared to $K_2 = 4$. Essentially, the most appropriate choice of $K_2$ is depended on the diversity of source domains. The more diverse the domains are, the more specific clusters are required to capture the specific cues. But it requires prior knowledges of the diversity, so we uniformly set $K_2 = 4$ for all the settings in the experiment

\subsection{Limited Source Domains}
We also evaluate the proposed method on the limited source domains setting where only two source domains are available. As shown in Table~\ref{tab:limited_domain}, the MSU-MFSD and Idiap replay-attack datasets are adopted as source domains and CASIA-FASD (M\&I to C) or OULU-NPU (M\&I to O) is the target domain. 
As can be seen, the proposed VLAD-VSA method surpasses most of existing works for two settings. The VLAD-VSA is comparable to the state-of-the-art MD-DRL \cite{wang20cross} on M\&I to C setting. For the M\&I to O setting, where the target domain has more videos than the two source domains, the improvement is more pronounced for the two metrics. It demonstrates that the proposed VLAD-VSA method is still effective to get generalized and discriminative representations in the challenging case with limited source domains and limited training data.

\begin{table}[]
	\centering
	\caption{Comparison of vocabulary separation strategy with different specific words. Vocabulary size $K = 32$.}
	\begin{tabular}{l|c|c|c|c}   
		\toprule
		\multirow{2}*{\textbf{Method}} & \multicolumn{2}{c|}{\textbf{O\&C\&I to M}} & \multicolumn{2}{c}{\textbf{O\&C\&M to I}}  \\
		\cline{2-5}
		~ & HTER (\%) & AUC(\%) & HTER(\%) & AUC(\%)\\
		\hline
		VLAD ($K_2 = 0$)  & 7.14 & 97.81 &15.64 & 93.72 \\
		+ VS ($K_2 = 2$) &  5.71 &97.66 & 15.00  &  91.23  \\
		+ VS ($K_2 = 4$) & \textbf{5.71} & \textbf{97.86} & \textbf{9.14} & \textbf{96.33}  \\
		+ VS ($K_2 = 6$) & 7.14 & 97.71 & 14.29 & 92.16 \\
		+ VS ($K_2 = 10$) & 8.57  & 96.47 &15.64   & 92.67\\
		\bottomrule
\end{tabular}
	\label{tab:sp_num}
\end{table}

\begin{table}[]
	\centering
	\caption{Comparison of cross-domain PAD results with limited source domains (MADDG).}
	\begin{tabular}{c|c|c|c|c}
		\toprule
		\multirow{2}*{\textbf{Method}} & \multicolumn{2}{c|}{\textbf{M\&I to C}} & \multicolumn{2}{c}{\textbf{M\&I to O}} \\
		\cline{2-5}
		~ & HTER & AUC & HTER & AUC \\
		\hline
		MS-LBP \cite{maatta2011face} &  51.16 & 52.09 & 43.63 & 58.07 \\
		IDA \cite{wen15face} & 45.16 & 58.80 & 54.52 & 42.17 \\
		CT \cite{boulkenafet2016faceLBP} & 55.17 & 46.89 & 53.31 & 45.16 \\
		LBP-TOP \cite{de2014face}  & 45.27 & 54.88 & 47.26 & 50.21 \\ 
		MADDG (M) \cite{shao2019multi} & 41.02 & 64.33 & 39.35 & 65.10 \\
		SSDG (M) \cite{jia20single} & 31.89  & 71.29 & 36.01 & 66.88 \\
		MD-DRL \cite{wang20cross} & 31.67 &\textbf{75.23} & 34.02 & 72.65 \\
		\hline
		VLAD-VSA (M) & \textbf{31.57}  &  74.03 &\textbf{26.39} & \textbf{80.67}
		\\\bottomrule
	\end{tabular}
	\label{tab:limited_domain}
\end{table}

\begin{figure*}
	\centerline{
		\subfigure[\normalsize $\lambda$ = 0]{
			\includegraphics[scale = 0.4,trim = 20 20 20 20,clip]{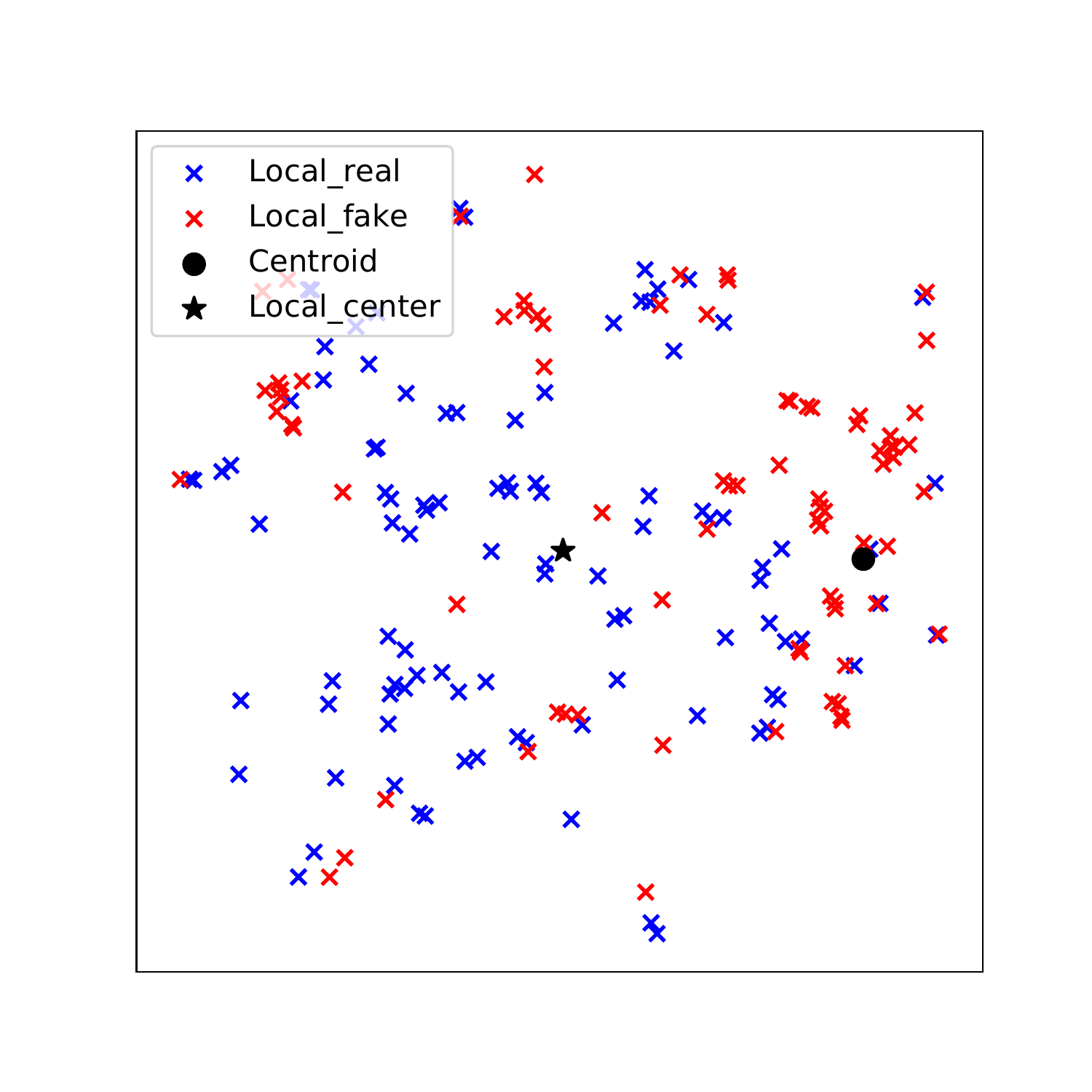}
		}
		\subfigure[\normalsize $\lambda$ = 0.1]{
			\includegraphics[scale = 0.4,trim = 20 20 20 20,clip]{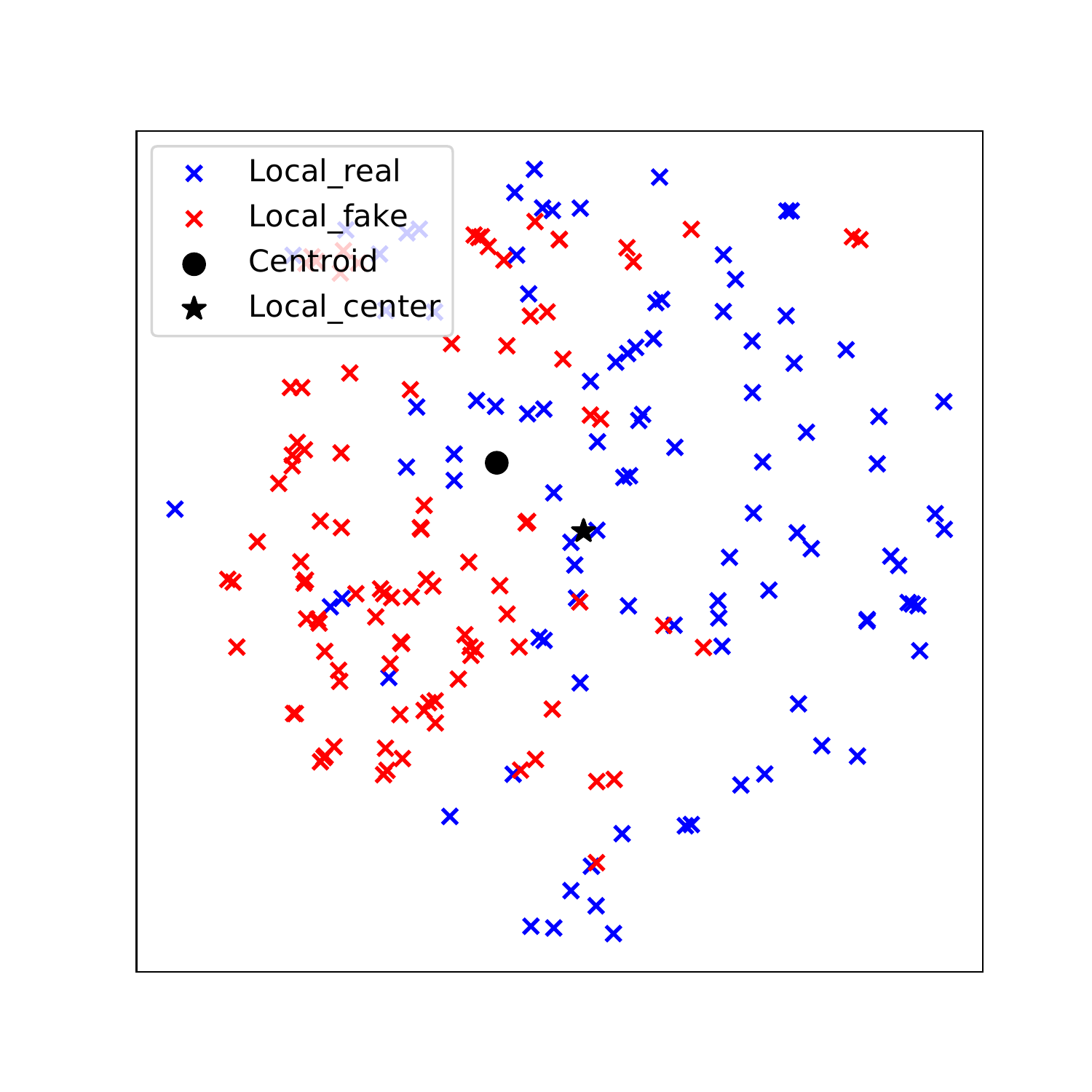}
		}
		\subfigure[\normalsize $\lambda$ = 1]{
			\includegraphics[scale = 0.4,trim = 20 20 20 20,clip]{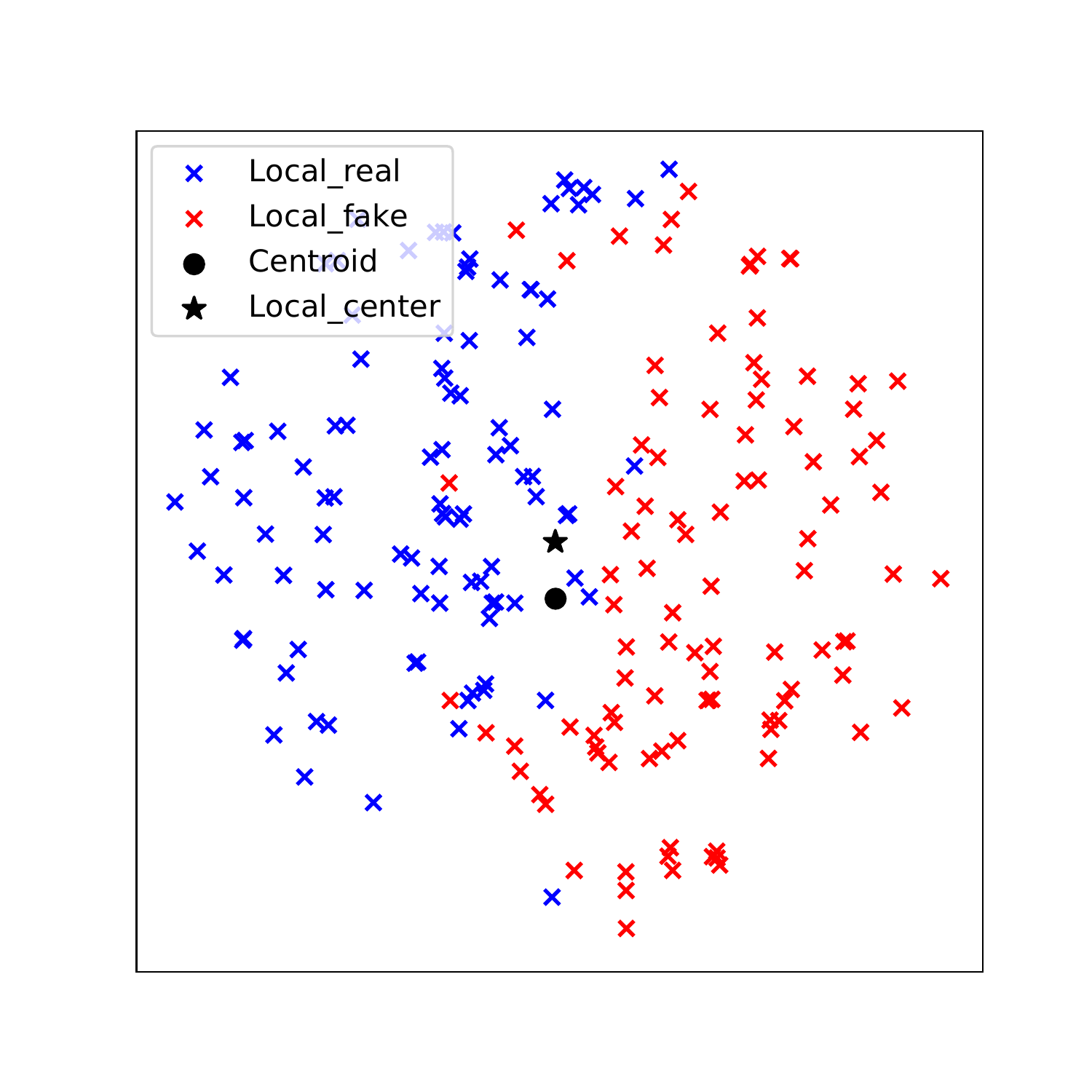}
		}
	}
	\caption{The T-SNE visualization of randomly sampled 100 real features, 100 fake features, feature center and cluster centroid in the second cluster. The distribution of them is impacted by the vocabulary adaptation method under different factor $\lambda$.}
	\label{fig:quali}
\end{figure*}

\begin{figure}
	\centerline{
		\subfigure[\normalsize Real / Fake]{
			\includegraphics[scale = 0.3,trim = 0 10 20 20,clip]{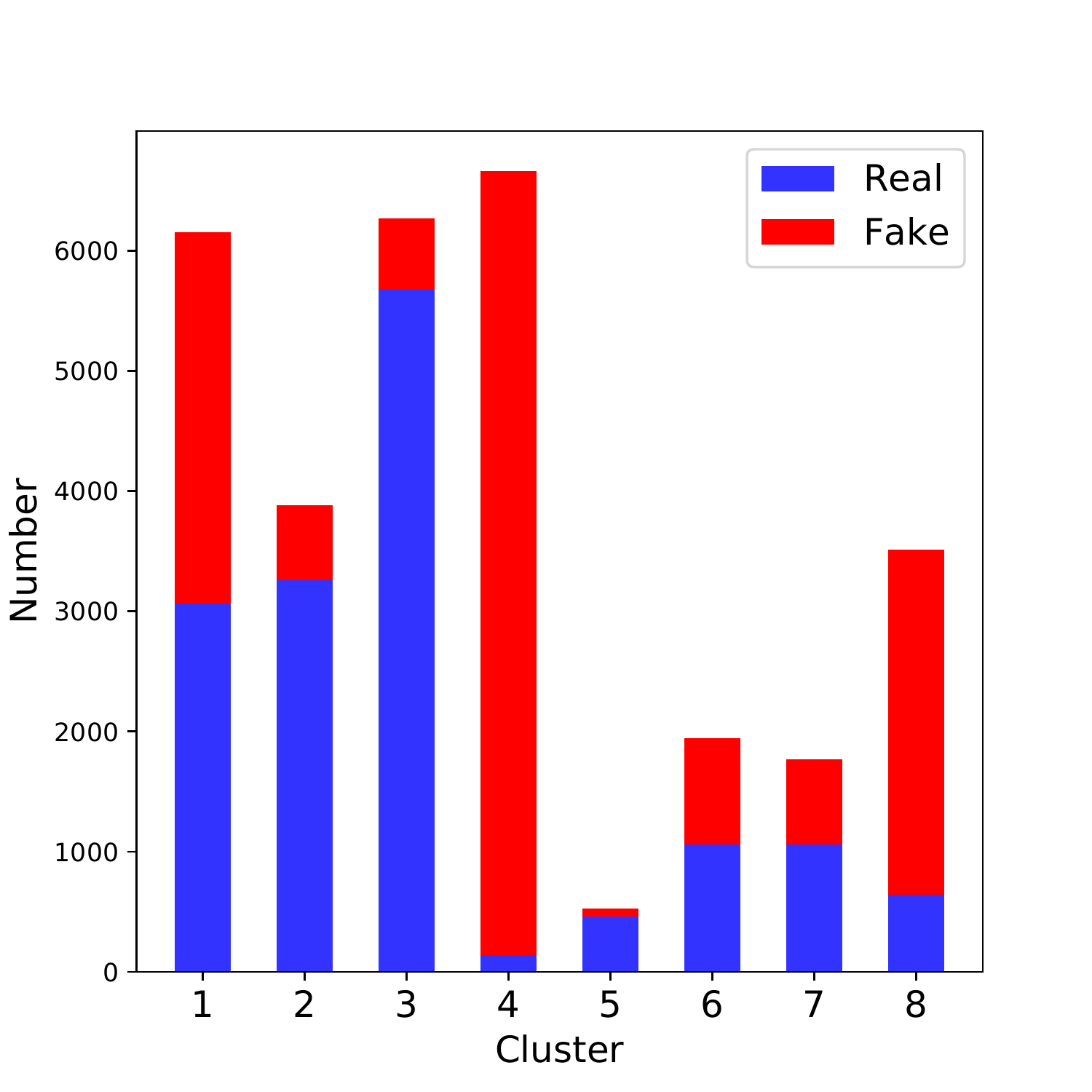}
		}
		\subfigure[\normalsize Domain]{
			\includegraphics[scale = 0.3,trim = 0 10 20 20,clip]{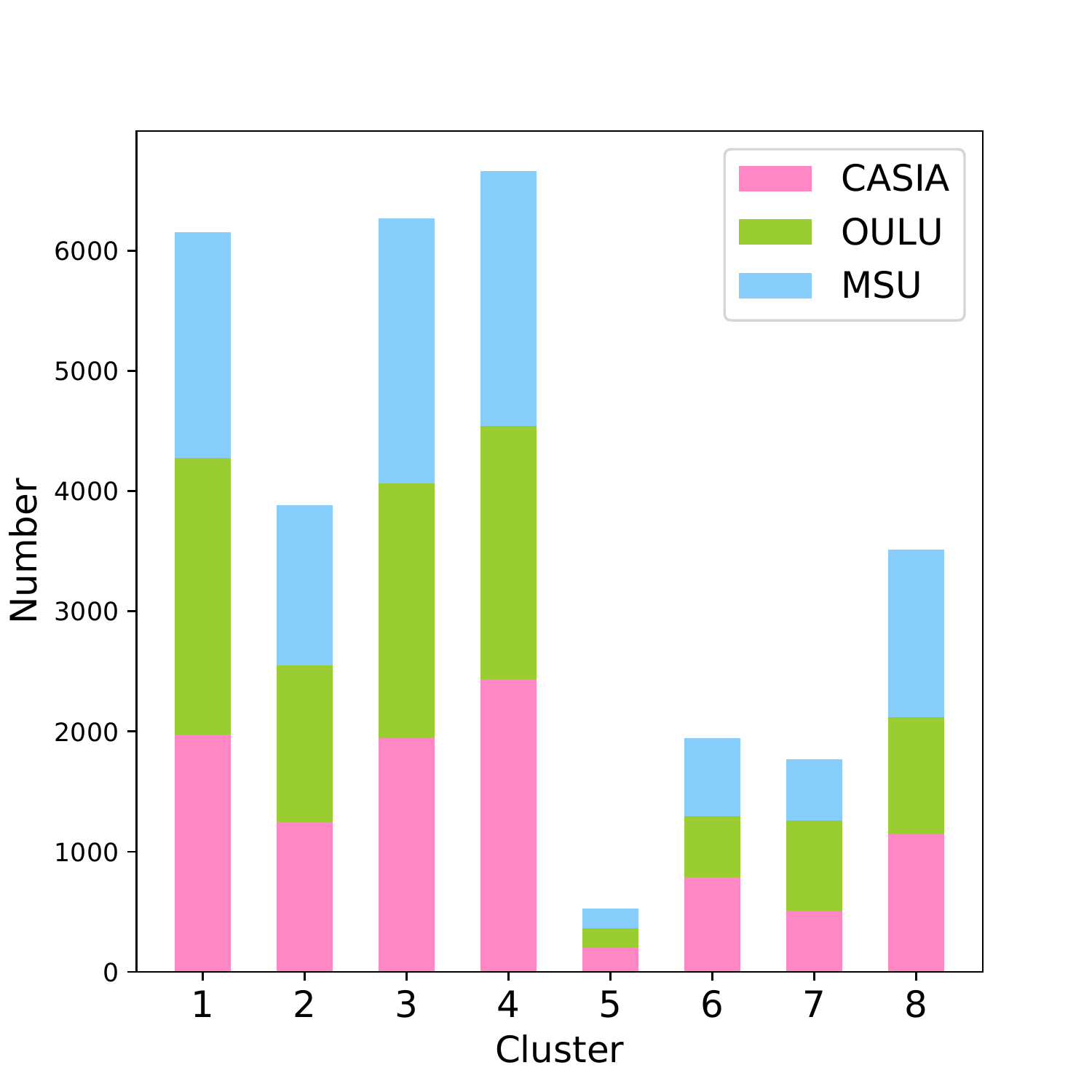}
		}
	}
	\caption{Visualization of the assignment statistics about local features to the clusters.} 
	\label{fig:cluster}
\end{figure}
\subsection{Qualitative Results} \label{sec:quali}

The proposed vocabulary adaptation method consists of centroid adaptation guarantees the cluster centroid be close to the assigned feature center and intra-cluster discriminative loss improves the intra-cluster discriminability.  To visually demonstrate their effectiveness, we adopt T-SNE to visualize the 2-D distribution of the local features and centroid of the second cluster in Figure~\ref{fig:quali}. 
Three sub-figures are shown with different magnitudes of $\lambda$, which control the influences of two losses on the feature distribution. 
As can be seen, the real and fake features in conventional VLAD ($\lambda=0$) are not so distinguishable, and the cluster centroid is far away from the local feature center. When gradually expanding the influence of two losses with ($\lambda=0.1, 1$), the real and fake features in the clusters get distinguishable and the centroid gets close to the local feature center to give reasonable residual calculation. In the training stage, we adopt $\lambda=0.1$ to complement other losses.

To understand the assignment status of local features to the visual clusters, we train a VLAD-VSA model with eight visual words (seven shared words and one specific word) on the OULU-NPU, CASIA-FASD and MSU-MFSD datasets. We randomly select 70 real and 70 fake images in each domain and count the local feature assignment statistics in the clusters. The assignment status of real-fake features and the multi-domain features is illustrated in Figure~\ref{fig:cluster},  from which there are three observations can be outlined: 
First, the feature number is not evenly distributed among clusters. It is natural that some features are usually more frequent than the others, and so to the visual words \cite{H2009On}.
Second, most of the visual words tend to capture particular real or fake cues.  As shown in Figure~\ref{fig:cluster} (a), the clusters are usually dominant with real or fake features. For example, most of the local features in fourth cluster are fake features and most of the local features in third cluster are real.  It demonstrates the visual words tend to capture the particular cues for recognizing real or fake faces. The specific (eighth) word is dominated with fake features, demonstrating the fake images are prone to contain the patterns specific in one dataset.
Third, the visual words are generalized across domains.  In Figure~\ref{fig:cluster} (b), we can find the clusters are usually domain-unrelated because the feature number of three domains is similar in all clusters. The clusters are not significantly biased to one domain, which demonstrates their generalization ability.

\section{Conclusion}
In this paper, we adopt the VLAD aggregation to preserve the local feature discriminability and get the discriminative representation. We modify VLAD with the proposed vocabulary separation and adaptation method for cross-domain PAD task. The VS method tackles the diversity among domains with shared-specific modeling. The VA method perfects the vocabulary optimization with centroid adaptation to guarantee robust similarity measurement and intra-cluster loss to explicitly constrain the intra-cluster discriminability. Extensive illustrations and experiments demonstrate the effectiveness of the proposed VLAD-VSA method, which gets new state-of-the-art results on the standard cross-domain PAD benchmarks. 

\section*{Acknowledgments}
This work was supported in part by the National Key R$\&$D Program of China under Grant No.2020YFC0832505, National Natural Science Foundation of China under Grant No.61836002, No.62072397 and Zhejiang Natural Science Foundation under Grant LR19F020006. Specially thank Yunpei Jia for sharing codes and insights.

\bibliographystyle{ACM-Reference-Format}
\balance
\bibliography{vlad}

\end{document}